%% file: ms.tex
\pdfoutput=1

\documentclass[letterpaper, 10pt, conference]{ieeeconf}      %

\IEEEoverridecommandlockouts                              %

\let\labelindent\relax                                    %

\usepackage{graphicx}            %
\usepackage{color}
\usepackage{amsmath}             %
\usepackage{amssymb}             %
\usepackage{amsfonts}            %
\usepackage{enumitem}            %
\usepackage{multirow}            %
\usepackage{siunitx}             %
\usepackage{url}                 %
\usepackage{xspace}              %
\usepackage[T1]{fontenc}         %
\usepackage[hidelinks]{hyperref} %
\usepackage{balance}
\usepackage[bottom]{footmisc}
\usepackage{xfrac}
\usepackage{flushend}
\usepackage{tensor}

\graphicspath{{images/}}
\DeclareGraphicsExtensions{.pdf,.png,.jpg,.jpeg}

\DeclareMathOperator{\sgn}{sgn}

\DeclareMathOperator{\acos}{acos}

\DeclareMathOperator{\atantwo}{atan2}

\providecommand{\norm}[1]{\lVert#1\rVert}

\newcommand{\degreem}{^{\circ}} %
\newcommand{\mypm}{\mathbin{\mathpalette\@mypm\relax}}

\newcommand{\seclabel}[1]{\label{sec:#1}}

\newcommand{\figlabel}[1]{\label{fig:#1}}

\newcommand{\eqnlabel}[1]{\label{eqn:#1}}

\newcommand{\secref}[1]{Section~\ref{sec:#1}\xspace}

\newcommand{\figref}[1]{Fig.~\ref{fig:#1}\xspace}

\newcommand{\iguhop}{igus\textsuperscript{\tiny\circledR}$\!$ Humanoid Open Platform\xspace}

\newcommand{\degree}{$\degreem$\xspace}

\title{\LARGE \textbf{Online Balanced Motion Generation for Humanoid Robots}}

\author{Grzegorz Ficht and Sven Behnke%
\thanks{All authors are with the Autonomous Intelligent Systems (AIS) Group, Computer Science Institute VI,
        University of Bonn, Germany. Email: {\tt\small ficht@ais.uni-bonn.de}. This work was partially
        funded by grant BE 2556/13 of the German Research Foundation (DFG).}}

\usepackage{eso-pic}

\AtBeginDocument{\AddToShipoutPictureFG*{\AtTextUpperLeft{\put(0,\LenToUnit{9pt}){\parbox{\textwidth}{\centering\bfseries
International Conference on Humanoid Robots (Humanoids), Beijing, China, 2018
}}}}}
\begin{document}

\bstctlcite{IEEEexample:BSTcontrol}

\maketitle
\thispagestyle{empty}
\pagestyle{empty}

\begin{abstract}
 Reducing the complexity of higher order problems can enable solving them in analytical ways. 
 In this paper, we propose an analytic whole body motion generator for humanoid robots.
 Our approach targets inexpensive
 platforms that possess position controlled joints and have limited feedback capabilities. By analysing the mass
 distribution in a humanoid-like body, we find relations between limb movement and their respective CoM 
 positions. A full pose of a humanoid robot is then described with five point-masses, with one attached
 to the trunk and the remaining four assigned to each limb. The weighted sum of these masses 
 in combination with a contact point form an inverted pendulum. We then generate statically stable poses by
 specifying a desired upright pendulum orientation, and any desired trunk orientation. Limb and trunk placement
 strategies are utilised to meet the reference CoM position. A set of these poses 
 is interpolated to achieve stable whole body motions. The approach is evaluated by performing several 
 motions with an \iguhop robot. We demonstrate the extendability of the approach by applying basic 
 feedback mechanisms for disturbance rejection and tracking error minimisation.
\end{abstract}
\section{Introduction}

For humanoid robots to be able to operate in a human environment, 
a wide array of motoric skills is necessary, where keeping
the robot balanced is of utmost importance. This is particularly difficult
due to the possible mechanical inaccuracies, inexact actuation, and inaccurate sensor feedback. 
The impact of these factors varies between robots, and is highly correlated 
with their price. Smaller, low-cost platforms, such as the widely used Aldebaran 
Nao, can cope better with these limitations---due to a comparatively high torque 
to weight ratio, which allows them to ignore several kinematic and dynamic constraints when performing an action. 
This is often not the case with larger-sized platforms possessing a lower torque to weight
ratio, which require more sophisticated control approaches. The current state-of-the-art however, focuses
mostly on expensive high-quality robots, which rely on high control loop
frequencies, precise torque-controlled actuators, and a considerable amount
of feedback information. Often, numerical optimisation methods are
used for their control. Simplifying the complex kinematic models
can enable analytic solutions.

In this paper, we present an analytic method for run-time generation
of parametric quasi-static motions from inherently balanced static pose keyframes,
based on the concept of triangle centroid mass (see \figref{triangle_teaser}),
The limbs and whole body of a robot are described with triangles, which
allows for finding a direct mapping between body movement and its respective CoM location.
The proposed method is applicable to larger robots, equipped with 
inexpensive actuators and sensors. By reducing the complexity of the 
model, we are able to make the robot balance in various configurations 
without any information about the forces or torques. We rely exclusively on
an on-board 6-axis IMU and joint positions.

\begin{figure}[!t]
\centering{\includegraphics[width=0.99\linewidth]{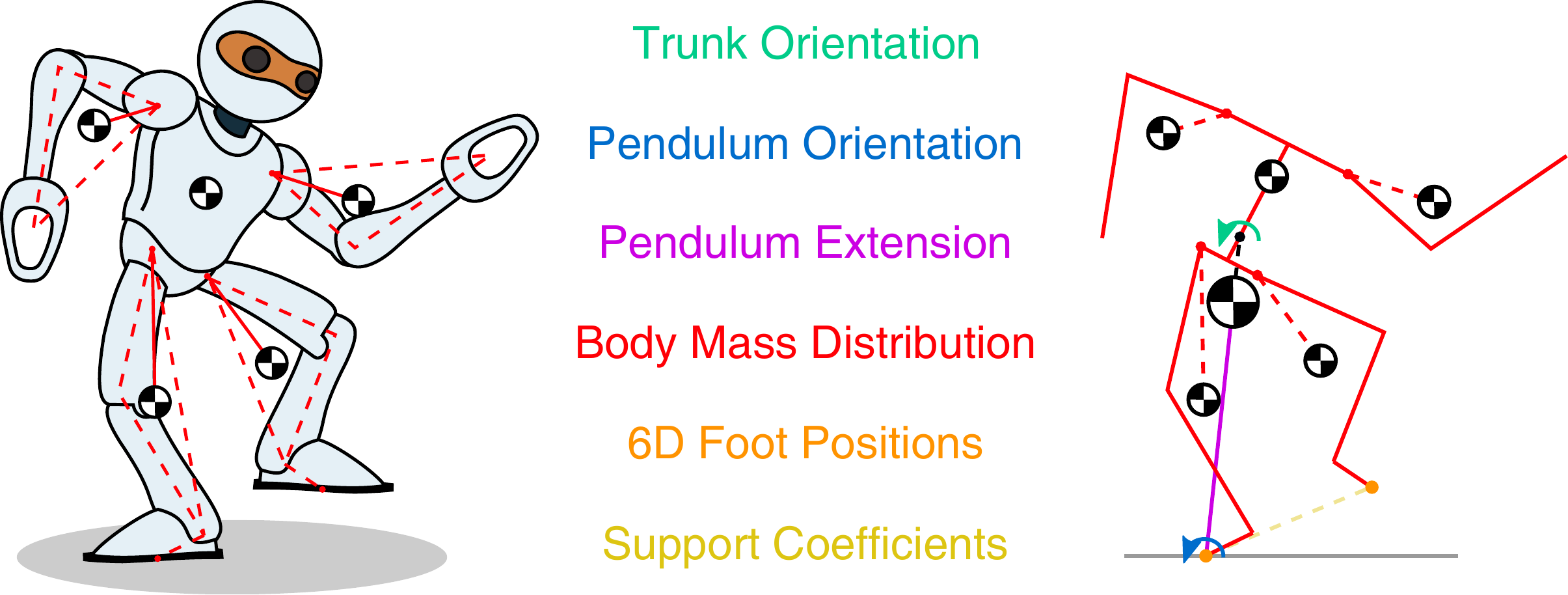}}
\caption{Approximation of a humanoid body and its limbs with triangles. The approximation allows to 
parameterise a whole body pose with only a few parameters to produce balanced motions.}
\figlabel{triangle_teaser}
\vspace{-2ex}
\end{figure}
\section{Related Work}

The first motion generation methods for humanoid robots pre-computed joint trajectories with little online 
modifications. Hirai et al.~\cite{hirai1998development} for example, derived a reference 
body trajectory based on the desired Zero Moment Point (ZMP)
and generated joint angle displacements to shift the ZMP to an 
appropriate position to maintain balance while walking with the 
Honda P2 platform. A different approach to this problem has 
been proposed by Fujimoto et al.~\cite{fujimoto2002simulation},
who perform the planning and control in a combined 
manner to track the reference Center of Mass (CoM) trajectory with 
respect to ground reaction forces in a real-time simulated environment. 
These works focus on solving the single task of generating motions
to keep a humanoid robot balanced while walking. 

Methods of generating whole-body motions for humanoids in a more general way have been 
developed by Kajita et al. in~\cite{kajita2003resolved}. 
Given reference linear and angular momenta while considering the constraints presented by contacts, a balanced 
kicking and walking motion was generated and performed by the HRP-2 robot
through joint velocity control. A different approach to the same problem
was developed by Sentis and Khatib~\cite{sentis2005synthesis}, 
where the control of lower priority tasks is projected into the 
null-space of the higher priority ones. The proposed hierarchical 
framework addresses a large set of constraints and provides compliant 
torque-based control that allows humanoid robots to simultaneously 
perform multiple tasks of varying complexity. These two 
motion generation techniques have led to the emergence of multiple 
Whole-Body Control (WBC) systems utilising joint velocities or 
torques to generate and perform motions. Position-controlled actuators
are not addressed by these control schemes. Although it is 
possible to integrate the results to achieve a desired position, due
to the accumulated noise, these approaches are hardly applicable on low-cost, real robots. 

The output produced by the Whole-Body Controller is only one way to categorise
them. Another way of distinction is by the method of finding a solution, 
which can be divided into closed-form and optimisation-based techniques. 
The former use a series of mathematical operations to achieve a desired 
joint trajectory, as in the mentioned approaches~\cite{kajita2003resolved, sentis2005synthesis}.
In the latter, the result is computed by a solver from a user-defined optimisation
problem, which is generally more time consuming. Many recent works~\cite{del2014prioritized, 
herzog2014balancing, feng2015optimization} are focused
on using solvers, as they allow for finding solutions to possibly conflicting tasks.
The long computation time, however often makes these types of methods
unfavorable when it comes to using them online on real robots, as they cannot
be run at interactive rates. To remedy this, Dai et al.~\cite{dai2014whole}
use the robot's full kinematics with reduced dynamics, as the full dynamics
did not allow the solver to converge even after days of computations. With
the proposed approach they are able to interactively generate motions in 
less than \SI{0.2}{s}. 

Reducing the complexity of a humanoid robot representation, while retaining most of 
the relevant information for control purposes led to the development of various models. Kajita et al.~\cite{kajita20013d}
introduced the 3D Linear Inverted Pendulum, which allowed for generation of a biped walk depending
on desired velocity and direction. In this model, the mass stays at a constant height and 
a change in the robot's configuration cannot influence it by definition. The Reaction Mass Pendulum 
described by Lee et al.~\cite{lee2007reaction} is a more comprehensive model, that uses three pairs 
of equal point masses at different radial distances to shape the inertia of the robot.
These masses are abstract however, and do not map directly to a specific limb or body part.
Describing the robot by distributed masses was also done by Takenaka et al.~\cite{takenaka2009real} 
for the purpose of fast gait generation. Their simplified model allows for fast calculations, but does not consider upper body movement and kinematic constraints.

Regardless of how the joint trajectories were generated, specification of
a stability criterion is necessary to keep the robot balanced during a motion. 
Most of the current developments make certain that the ZMP lies inside the support
polygon, be it the original definition~\cite{vukobratovic2004zero}, or a more generalised 
one~\cite{harada2003zmp}. A universal method of assessing stability was also 
presented by Hirukawa et al. in~\cite{hirukawa2006universal}, where stability 
is judged by whether the gravity and inertia wrench lies inside the polyhedral 
convex cone of the contact wrench between the robot and its environment. 

To properly evaluate the stability through e.g. the ZMP, Whole-Body 
Control techniques rely on good sensor feedback. This limits their usability 
to higher-quality hardware where force-torque sensors are available.  
The ZMP could also be estimated through the use of inverse dynamics,
but that again requires accurate feedback data as even small noise can 
greatly influence the result. These factors are frequently an issue
with lower cost platforms, and in combination with an unfavorable torque to 
weight ratio, the applicability of WBC on them is highly restricted. %

\section{Model Complexity Reduction} \seclabel{model_reduction}

When considering a humanoid body plan, the CoM is not fixed to any of the links, but rather moves in a 
certain region depending on the placement of body parts. This region lies in the vicinity of the lower part of the trunk---as it is the heaviest part of the 
body with 50 to 55 percent of the total mass, following with the legs (from 30 to 35 percent)~\cite{drillis1966body}. 
In any standing pose scenario, the legs perform a supporting function 
by connecting the hipline to the contacting surface (ground). If the contact is approximated by a single origin point $O$, the spine is stiff and begins at the
hipline midpoint $H$ and ends at the shoulder midpoint $S$, then it is possible to describe such a pose by a triangle connecting these points $\triangle OSH$.
If the body is symmetrical, the resulting CoM will lie inside of this triangle, with its exact position depending on the
mass distribution between all body parts. This position can be considered as nominal and(depending on limb movement) can still change.
The contact point and CoM form a pendulum that can be used to describe the stability of the robot.
We consider the neck and head as a part of the trunk, as the limited motion it can achieve paired with the relatively low weight
does not influence the CoM in a significant way. Furthermore, unconstrained head movement is desirable for the vision system. 
These assumptions are the basis of the work described in this section.

Other than a humanoid's whole body pose being represented by a triangle, the same characteristic can be
observed in the arms and legs~(\figref{triangle_teaser}). By placing points at the hips $H^L,H^R$, 
knees $K^L,K^R$ and ankles $A^L,A^R$ for the legs
as well as the shoulders $S^L,S^R$, elbows $E^L,E^R$, and wrists $W^L,W^R$ for the arms, 
four triangles can be defined with an accompanying CoM that lies within them.
These limb triangles have one side with variable length, depending on the extension of 
the limb and two sides with known and fixed lengths.
In the arms, these are the upper arm and forearm while in the leg they are represented by
the thigh and shank. By parameterising the mass distribution in the limbs, it is possible to associate
joint angles and the limb mass placement in space with a direct mapping~(see \secref{centroidal_mass}). As with the head and 
the trunk, we assume the feet and hands do not influence the CoM of their respective 
limb---due to their limited range of movement and comparatively low weight.

With these assumptions, it is possible to describe a pose of a humanoid robot through the use of five point masses, where 
the main one is attached to the torso, and the remaining four are assigned to the limbs.
The trunks centre of mass is described with a three-dimensional offset from the shoulder midpoint.
The origins of the limbs are located at the shoulders and hips, which are spaced symmetrically 
away from their respective midpoints at exactly half the shoulder width $s_w$ and hip width $h_w$, accordingly.
The whole body mass is a weighted sum of the five body part mass coordinates, and forms a pendulum
with the single ground contact point. The proposed approximation might not provide any benefit when 
doing a forward calculation of the CoM based on joint positions and physical properties of the links. The inverse operation 
of placing the robots CoM in space however, is greatly simplified and allows to utilise analytical limb mass 
placement strategies that do not involve optimisation. Generating a pose based on the desired CoM position is the 
main contribution of the presented work and is described in detail in \secref{pose_generation}. 
Due to the assumptions made, it is possible that some accuracy with regard to
CoM tracking may be lost. In \secref{results} we demonstrate that this effect is minor.

\subsection{Centroidal mass in a triangle} \seclabel{centroidal_mass}

Given a triangle with uniform density, the centroid is 
located at the intersection of its three medians, however only two of 
them are necessary to determine it. The centroid divides each of the 
medians with a ratio of 2:1, meaning that it is located \sfrac{2}{3} of 
the distance from any selected vertex to the midpoint of the side opposing it.
The location of the mass center can therefore be described with as 
little as two values: $p_s$ and $p_l$---named the CoM distribution parameters.
$p_s$ is the ratio of chosen side $s$ at which the line segment $l$ from the 
opposing vertex $v$ sections it to its length, and $p_l$ is the 
ratio of the distance from the same vertex $v$ along line $l$ at which the CoM 
is located to that line's length. In a regular triangle $p_s = \frac{1}{2}$, while 
$p_l = \frac{2}{3}$, and line $l$ is a median. By manipulating these parameters, 
it is possible to place the CoM anywhere inside the triangle in accordance 
to its mass distribution.

Let us consider an example in two-dimensional space to depict the possible solution.
Given a desired CoM position $M$ in regards to the origin $A$ 
with two known sides $a,c$ and the CoM distribution parameters $p_s,p_l$, compute the 
missing angles $\alpha, \beta, \gamma$ and the length of side $b$ in triangle $\triangle ABC$ as shown on \figref{triangle_mass}. 
The line segment $l$ dividing $\triangle ABC$ at point $P$ into two smaller triangles $\triangle ABP, \triangle APC$
can be computed as
\begin{figure}[!t]
\centering
\def\svgwidth{0.8\linewidth}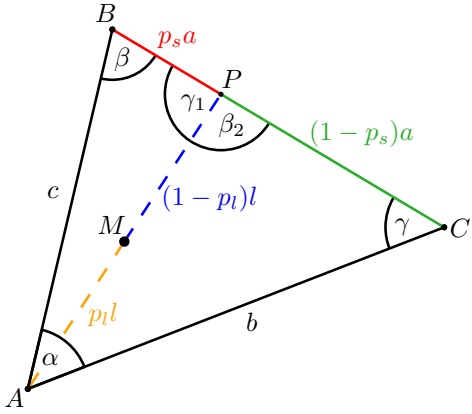
\caption{Mass distribution in a triangle. Knowing the distribution parameters $p_s$,$p_l$ and two sides $a$,$c$, it is possible
to calculate all of the angles, given a desired mass placement $M$ in respect to origin $A$.}
\figlabel{triangle_mass}
\vspace{-3ex}
\end{figure}
\begin{equation}
l = d(A,M)/p_l, \eqnlabel{length}
\end{equation}
where $d(A,M)$ is the Euclidean distance between points $A$ and $M$. 
From the law of cosines, we compute the relations in triangles $\triangle ABP, \triangle APC$:
\begin{align}
c^2 &= l^2 + (p_s a)^2 - 2 p_s l \cos(\gamma_{1}), \label{triangleEqStart}   \\
b^2 &= l^2 + ((1-p_s)a)^2 - 2 (1-p_s) l \cos(\pi - \gamma_{1}).
\end{align}
For a cleaner notation let us introduce a complementary ratio $p_c = 1-p_s$, and knowing that 
$\cos(\pi - \gamma_{1}) = -\cos(\gamma_{1})$, we can solve for $b$:
\begin{equation}
b = \sqrt{ 2 l^2 + p_c^2 a^2 + p_s^2 a^2 +2(p_c-p_s) l a \cos(\gamma_{1}) - c^2}.
\end{equation}
Angle $\gamma_{1}$ in triangle $\triangle ABP$ can be computed as
\begin{align} 
\alpha_{1} & = \acos(\frac{-p_s^2 a^2 + l^2 + c^2} {2 l c} \label{alpha}), \\ 
\beta_{1} & = \acos(\frac{p_s^2 a^2 - l^2 + c^2} {2 p_s a c}\label{beta}), \\ 
\gamma_{1} & = \pi - \alpha_{1} - \beta_{1} \label{gamma}.
\end{align}
Finally, having all side lengths of triangle $\triangle ABC$, we compute its $\alpha,\beta,\gamma$ angles
similarly as in Eqs. (\ref{alpha}) to (\ref{gamma}), by substituting $b$ for $l$ and $a$ for $p_s a$.
The triangle describing the human pose in relation to the CoM of the whole body can be calculated 
in a similar manner as the one for the CoM of the limb with minor changes to the method. The difference 
stems from the fact that the triangle associated with the body has only a single side with invariable length---the spine length.
However, by knowing the orientation of the spine it is possible to compute its relative 
orientation $\gamma_{1}$ to the line segment $l$, which in turn allows to calculate the missing 
variables with the method above. This strongly relates to the trunk placement strategy, described in \secref{trunk_placement}.

\subsection{Projected angles}

\begin{figure}[!t]
\parbox{\linewidth}{\centering \def\svgwidth{0.85\linewidth}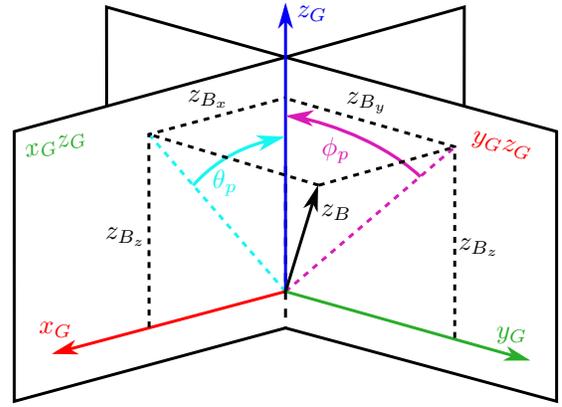}
\vspace{0.8ex}
\caption{Projected angles definition. A combination of two 2D angles $\theta_p$,$\phi_p$ on perpendicular planes $x_Gz_G$,$y_Gz_G$ is used to define a 3D orientation of a body's z-axis $z_B$.}
\figlabel{proj_angles}
\vspace{-3ex}
\end{figure}

We extend the two-dimensional approach to three dimensions with the use of projected angles. Many parameterisations 
for rotations exist with their inherent properties depending on the definition as shown in~\cite{Allgeuer2015}, where
the authors perform a review and summary of these and introduce two new representations---tilt angles and fused angles.
In the same work, it is mentioned that new parameterisations are created to provide convenient exploits that increase 
the efficiency of algorithms or ease the geometric interpretation. We motivate the usage of projected angles with the 
need to set the inclination of objects when projected onto separate global $xz$ and $yz$ planes, in a way where
an inclination on one plane does not influence the other. In this work, we assume a right-handed coordinate system, 
with the x-axis pointing forward, y-axis pointing to the left of the robots origin and z-axis pointing upwards.

As with fused angles, we base our definition on the intermediate tilt angles representation, 
and introduce the \textit{projected pitch} and \textit{projected roll} angles. The definition of
these projected angles is illustrated on \figref{proj_angles}. In fused angles, the $z_G$ vector is projected on the 
$y_B z_B$ and $x_B z_B$ planes which are fixed to the body, while projected angles perform the inverse operation
of projecting the $z_B$ vector onto the global $y_G z_G$ and $x_G z_G$ planes, which define the projected
pitch $\theta_p$ and projected roll $\phi_p$, respectively. The mathematical definitions are as follows:
\begin{align} 
\theta_p &= \atantwo(z_{Bx},z_{Bz}) &\in [-\pi,\pi], \\ 
\phi_p &= \atantwo(z_{By},z_{Bz}) &\in [-\pi,\pi].
\end{align}
The definition is completed with the fused yaw $\psi$, which performs the rotation around $z_B$, as in~\cite{Allgeuer2015}. 
It can be observed that projected angles are nonlinear and have a singularity when at least one of the angles is equal to $-\frac{\pi}{2}$ or $\frac{\pi}{2}$. 
To accommodate for these problems, we use spherical interpolation in combination with a prohibition band of negligible 
width around the singularity values.

The main advantage of projected angles is the ability to specify the global orientation of objects where 
a projected inclination remains the same, regardless of the projection in the plane perpendicular to it. 
This is not the case when performing the rotations sequentially, or with fused angles. 
Another benefit of this representation is that specifying a desired global orientation of the spine 
in combination with a set pendulum orientation makes the computation of $\gamma_1$ in Eq. (\ref{gamma})
straightforward in both the pitch and roll directions. The 3D solution is then equivalent to a pair 
of 2D solutions, greatly simplifying the calculations.

\section{Balanced motion generation}  \seclabel{pose_generation}

With the model description introduced in \secref{model_reduction}, 
we define a static whole body pose of a humanoid robot through a set of four components:
\begin{itemize}
 \item \textbf{Body pendulum:} projected roll $\phi_{p}^B$, projected pitch 
 $\theta_{p}^B$, heading $\omega^B$ and length $l^B$ encode the balance of a 
 humanoid robot into a pendulum with mass $m^B$.
 \item \textbf{Trunk orientation:} projected roll $\phi_{p}^T$, projected pitch 
 $\theta_{p}^T$ and fused yaw $\psi^T$ are responsible for orienting the trunk with a spine of length $l^T$ in 3D.
 \item \textbf{Feet positions:} 6D poses $F_l$, $F_r$ define the placement of the feet.
 \item \textbf{Support coefficient:} $c_s$ describes the weight distribution between the feet. 
\end{itemize}
The first step requires to determine the origin of the body pendulum $O$. 
This is done through a combination of the desired foot positions and the support coefficient.
The support coefficient represents the whole weight distribution between the legs.
Shifting the whole weight from the right to the left foot is expressed by $c_s \in [0,1]$---a 
value of $c_s = \frac{1}{2}$ means that both feet equally support the mass of the robot.
This is equivalent to moving the body pendulum origin between the left and right foot contact points.

Once the origin point has been determined, we utilise the desired pendulum orientation and length
to calculate the target CoM position:
\begin{equation} 
com^B = z^B  l^B + O,  \label{wbcom}
\end{equation}
where $z^B$ is a unit vector representing the z-axis direction of the body pendulum, 
calculated as
\begin{align} 
z^B_x &= \tan(\sgn(\cos(\theta_{p}^B))\theta_{p}^B)  \\ 
z^B_y &= -\tan(\sgn(\cos(\phi_{p}^B))\phi_{p}^B)  \\ 
z^B_z &= \sgn(\cos(\theta_{p}^B))
\end{align}
This method can be used to calculate any point in 3D space with the use of projected angles.
Through controlling the extension and orientation of the CoM similarly to a pendulum, we 
have the possibility to influence the stability of the robot.
In our approach, we constrain the body pendulum to an upright orientation to generate inherently balanced static poses.

\subsection{Trunk placement} \seclabel{trunk_placement}

The description of a humanoid robot pose through a triangle ties
the orientation of the trunk with the state of the body pendulum. As mentioned in 
\secref{centroidal_mass}, the line connecting the robot's origin with the total CoM
extends further and intersects the spine. This intersection point is called the 
trunk pivot point $t_{pp}$ and defines the origin of rotation for the trunk. It's coordinates
can be calculated as in~(\ref{wbcom}), by substituting the length with ${l^B}/{p_l^B}$. 
By defining a desired trunk orientation in the pose, we compute the hip and shoulder origins through
\begin{equation} 
\begin{bmatrix}
H^L_{xyz} \\
H^R_{xyz} \\
S^L_{xyz} \\
S^R_{xyz}
\end{bmatrix} 
= 
\begin{bmatrix}
0 & \frac{h_w}{2} & -p^B_s l^T \\
0 & -\frac{h_w}{2} & -p^B_s l^T \\
0 & \frac{s_w}{2} & (1-p^B_s) l^T \\
0 & -\frac{s_w}{2} & (1-p^B_s) l^T 
\end{bmatrix} 
R_T^{-1}+ t_{pp},
\end{equation}
where $R_T$ is the rotation matrix of the trunk, calculated with~\cite{MatOctRotLibGithub} as per~\cite{Allgeuer2015}.
The achieved point coordinates are then rotated around the global z-axis, at the pendulums origin by the heading value $\omega^G$.

\subsection{Foot placement}

The foot support points are offset from their respective ankles, and are 
located at the geometric centre of the feet. Given a desired 6D foot pose and 
knowing the foot offsets, we can directly compute the desired ankle position. 
Although the distance between the hip and ankle positions is sufficient to compute the 
extension of the leg CoM from its origin, it does not provide full
information for placing the CoM in 3D. For a full definition of the CoM placement, the 
leg rotation needs to be specified. To simplify the calculations, we assume 
that the ankle does not possess a yaw joint, and therefore the whole leg yaws 
at the hip. This is the case for most of the robotic platforms available currently on 
the market. This assumption is not necessary, but its lack would require an alternative 
method to specify the set leg yaw. 

We compute the leg yaw as the angle $\psi^L $ around the leg's z-axis $n_1$ between a set 
of planes which normals represent the base $n_2$ and desired $n_3$ pointing directions 
of the leg's x-axis. 
A unit vector created from the line segment connecting the hip 
to the ankle in combination with a fused yaw of zero is used to compute the $n_1$ z-axis 
and $n_2$ base x-axis direction. The desired x-axis direction vector $n_3$ is given by
\begin{equation}
n_3 = \frac{ n_1 \times n_y }{ \norm{n_1 \times n_y } },
\end{equation}
where $n_y$ is the normal of a plane passing through three points: the hip origin, 
desired ankle position, and an $xy$ offset from that position representing
the desired yaw component~(e.g. foot's x-axis direction vector, knee point etc.). 
The final leg yaw is computed as 
\begin{equation}
\psi^L = \pi - \atantwo((n_2 \times n_3) \cdot n_1, n_3 \cdot n_2 ),
\end{equation}
which allows to fully define the leg CoM position.

\subsection{Arm placement}

The arms play a vital role in balancing a humanoid robot. When walking or 
performing a wide array of motions (e.g.~kicking a ball, jumping), the arms 
can correct for a certain amount of error between the desired and current CoM 
at any given time. This is also partially true when one of the arms has an 
assigned task (e.g. carrying or reaching an object) as the other arm can 
move freely. Its relatively low weight can however mitigate only slight CoM 
displacements. Assuming that the legs~(LL, RL) and trunk~(T)
have been placed, a task has been assigned to the right arm~(RA) and its position
has been calculated in a similar manner to the legs, the left arm~(LA) desired 
CoM position then is simply
\begin{equation} \label{arm_mass}
 com^{LA} = \frac{com^B_w - (com^T_w + com^{LL}_w + com^{RL}_w + com^{RA}_w)}{m^{LA}}
\end{equation}
where $com_w$ are the mass-weighted CoM coordinates. The distance between $S^L$ and
$com^{LA}$ can be out of the movement range defined by the minimum $e_{min}$ and maximum $e_{max}$
CoM extension and needs to be altered to account for that. Following this,
the desired robot CoM placement cannot be met and requires a corrective
action from the trunk to satisfy it.

This claim still holds when both arms are free to move, accordingly with
the possible range of movement and their relative mass. With two arms however,
both of these parameters are greater, therefore the corrective actions have a 
stronger influence on the effective CoM position. In order to place both arms 
in a way that their single combined CoM satisfies the desired whole-body CoM position 
requirement, an arm placement strategy is needed. Naturally, this problem could 
be solved using optimisation with a set of constraints however, that would 
greatly increase the computation time. Knowing that in some cases the CoM displacement 
error cannot be brought to zero without modifying the trunk position, brings 
the problem back to that of whole body motion planning.

The proposed arm positioning strategy needs to provide a continuous solution comprised
of two coordinates $com^{RA}$,$com^{LA}$ that satisfies the desired CoM placement $com^A_d$ 
when possible, in respect to the arm's range of motion. Due to the symmetry of the whole body
in the $xz$ plane, the obvious approach would involve placing $com^{RA}$,$com^{LA}$ 
equidistantly on a line that crosses $com^A_d$ and is parallel to the trunks y-axis. 
The solution performs well only when the $y$ component of $com^A_d$ relative to the trunk
is close to zero, as the symmetrical motion of the arms counterbalances itself when hitting limits. 
In addition, this strategy requires that both arms move with the shifting of $com^A$, 
which produces unnatural looking poses. We slightly modify this method by using a 
different direction vector of the line supposed to cross $com^A_d$. For this, 
two shoulder spheres are created, that originate at $S^L$,$S^R$ and have a radius 
of $e_{max}$. The third sphere is placed at $com^A_d$, and represents the proximity 
of the solution to the shoulders, with a radius of
\begin{equation}
r = \min(d(com^A_d,S^L),d(com^A_d,S^R))+e_{max}.
\end{equation}
The two intersection $i^L$,$i^R$ points between the proximity sphere and the shoulder spheres
form the initial direction line, which is then shifted to cross $com^A_d$. 
In case the proximity sphere does not intersect the further shoulder sphere, 
the second intersection point is substituted by the point on that sphere's surface
nearest to $com^A_d$.
The final direction line is then adjusted to account for $e_{max}$ and a minimum $y$ 
distance of the arms from the trunk to avoid self-collisions. The strategy used 
produces natural looking poses. As the movement of both arms
is bound to the direction line which moves with $com^A_d$ and with respect to arm extension 
limits, the produced result is continuous.

\subsection{Performing motions}

A set of keyframes representing statically stable poses can be used to generate
stable motions, when the influence of dynamics is negligible, which is the case
when the motion is performed slowly.
For this purpose, based on the motion duration and control frequency we linearly 
interpolate all of the pose parameters apart from the projected pitch $\theta_{p}$ and 
roll $\phi_{p}$. Due to their intrinsic nonlinearity, we compute pairs of 
intermediate projected angles from the interpolated z-axis coordinates.
These coordinates lay on an arc of a great circle connecting the start $z^S$ and 
end $z^E$ pose z-axis orientation unit vector. The 
angle between $z^S$ and $z^E$ coordinates laying on the great circle plane
\begin{equation}
 \angle_{gc} = \atantwo( \norm{ z^S \times z^E }, z^S \cdot z^E)
\end{equation}
is linearly interpolated and used to rotate $z^S$ coordinates around the 
great circle plane normal $n_{gc}$. Care is also taken 
as to not wrap around $\pm \pi$ to avoid self-collisions and unwanted behaviour. 
In the majority of cases the result is a path on a minor arc which ensures minimum travel
between the two pose orientations.

\section{Calibration}

\begin{figure}[!t]
\parbox{\linewidth}{\centering\includegraphics[width=0.99\linewidth]{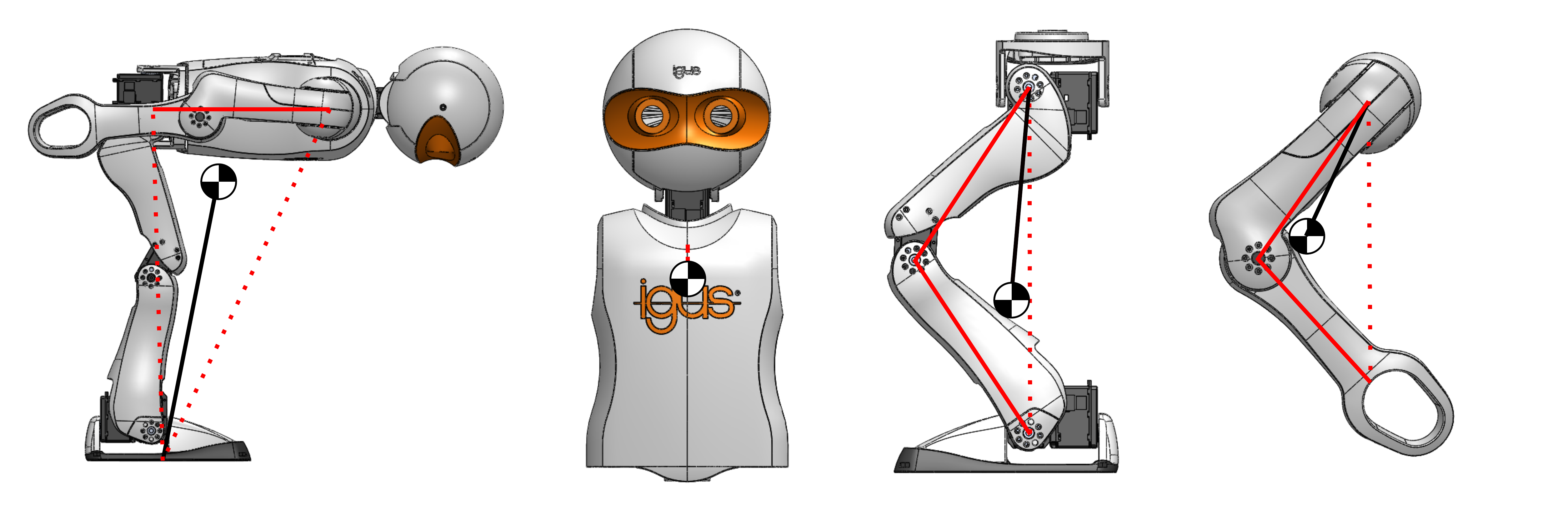}}
\caption{Mass distribution in a CAD model of the \iguhop. From left to right: whole body, trunk, leg, and arm.}
\figlabel{body_calib}
\vspace{-3ex}
\end{figure}

In order to use the method on a robot, a full body 
calibration to extract the $p_s$ and $p_l$ mass distribution parameters, $o^T_{xyz}$ trunk CoM offset 
and several body length measurements is necessary. For this purpose, a complete CAD model of 
the \iguhop robot was used~\cite{allgeuer2015child}. The measurements, masses and CoM positions for each of the five 
body parts, as well as the complete robot's CoM position have been extracted from the CAD data and used 
without alteration.

In the \iguhop, the shoulders are \SI{0.245}{m} apart, while the distance between the hips is \SI{0.11}{m}.
The spine length, connecting the shoulder and hipline midpoints equals \SI{0.225}{m}. The limb's 
upper and lower links are symmetrical and possess lengths of \SI{0.1995}{m} for the legs and \SI{0.17}{m} for the arms.
The geometric centers of the left and right foot are offset from their respective ankles by 
\begin{align} 
o^{LF}_{xyz} &= \begin{pmatrix}0.035 & -0.011 & -0.038 \end{pmatrix},  \\ 
o^{RF}_{xyz} &= \begin{pmatrix}0.035 &  0.011 & -0.038 \end{pmatrix} 
\end{align}
when the desired orientation of the foot remains vertical, and are subject to change with 
alterations to the foot orientation.

To retrieve the mass distribution information, we bent all of the limbs and the whole 
body at a right angle at their main joints. A \SI{90}{\degree} angle formed at the hips, elbows and knees
represents exactly half of the possible CoM extension, meaning that the calibration 
should provide equally good values for both $p_s$ and $p_l$. Calculating these parameters
is straightforward, given the full limb triangle description and relative CoM position.
An approximation of the resulting mass distribution is shown on \figref{body_calib}.

Although averaging multiple measurements with different arrangements can be done to 
achieve more accurate results, we have found that a single calibration with a right 
angle formed by the fixed sides of the limbs yields a good approximation for $p_s$ and $p_l$. 
The trunk mass of \SI{2.373}{kg} is offset from the shoulder midpoint by
\begin{equation}
 o^T_{xyz} = \begin{pmatrix}0.004 & 0 & -0.0475\end{pmatrix}.
\end{equation}

Due to the symmetry of the robot, the left and right limbs have the same mass and 
weight distribution. One arm weighs \SI{0.554}{kg}, while a leg has a mass of \SI{1.332}{kg}. 
The mass distribution parameters for the body, legs and arms used in our experiments are as follows:

\begin{equation} \label{mass_parameters}
\begin{pmatrix}
p^B_l p^B_s \\
p^L_l p^L_s \\
p^A_l p^A_s
\end{pmatrix} 
= 
\begin{pmatrix}
0.7743 & 0.5479 \\
0.6159 & 0.7930 \\
0.5608 & 0.1664 
\end{pmatrix}.
\end{equation}

\section{Experimental Results} \seclabel{results}

Verification of the proposed method was carried out with an \iguhop robot.
First, we evaluate the accuracy of the proposed approximation with the error between the CoM 
positions computed with our approach and the Universal Robotic Description Format (URDF) model for various poses of the robot.
Next, we test the capability of performing whole body posing with single and double leg
support poses, which we follow with performing a kicking motion.
The achieved results are then quantified by the CoM tracking error.

\subsection{Approximation error}

\begin{figure}[!t]
\parbox{\linewidth}{\centering
\includegraphics[width=0.49\linewidth]{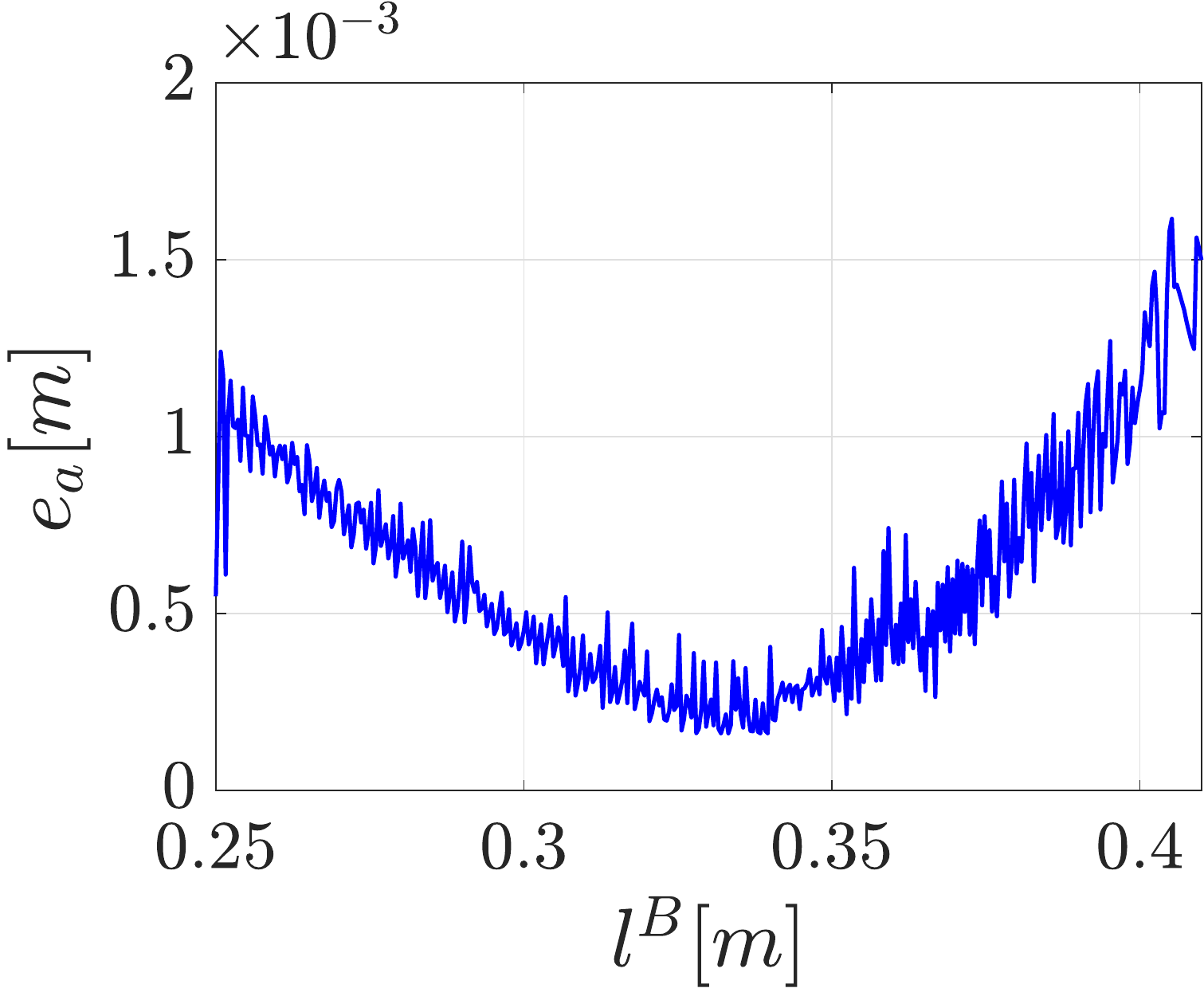}
\includegraphics[width=0.49\linewidth]{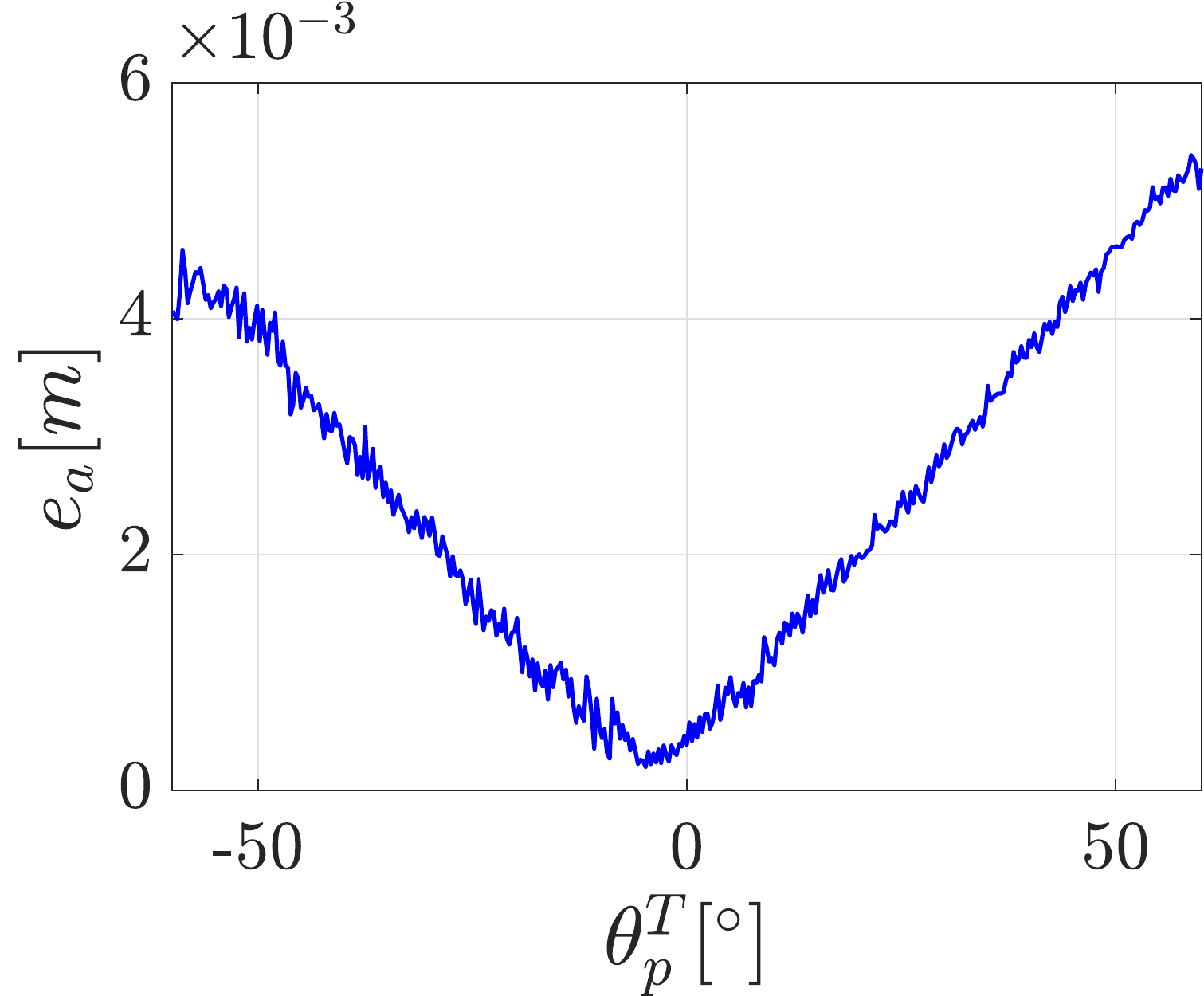}
\includegraphics[width=0.49\linewidth]{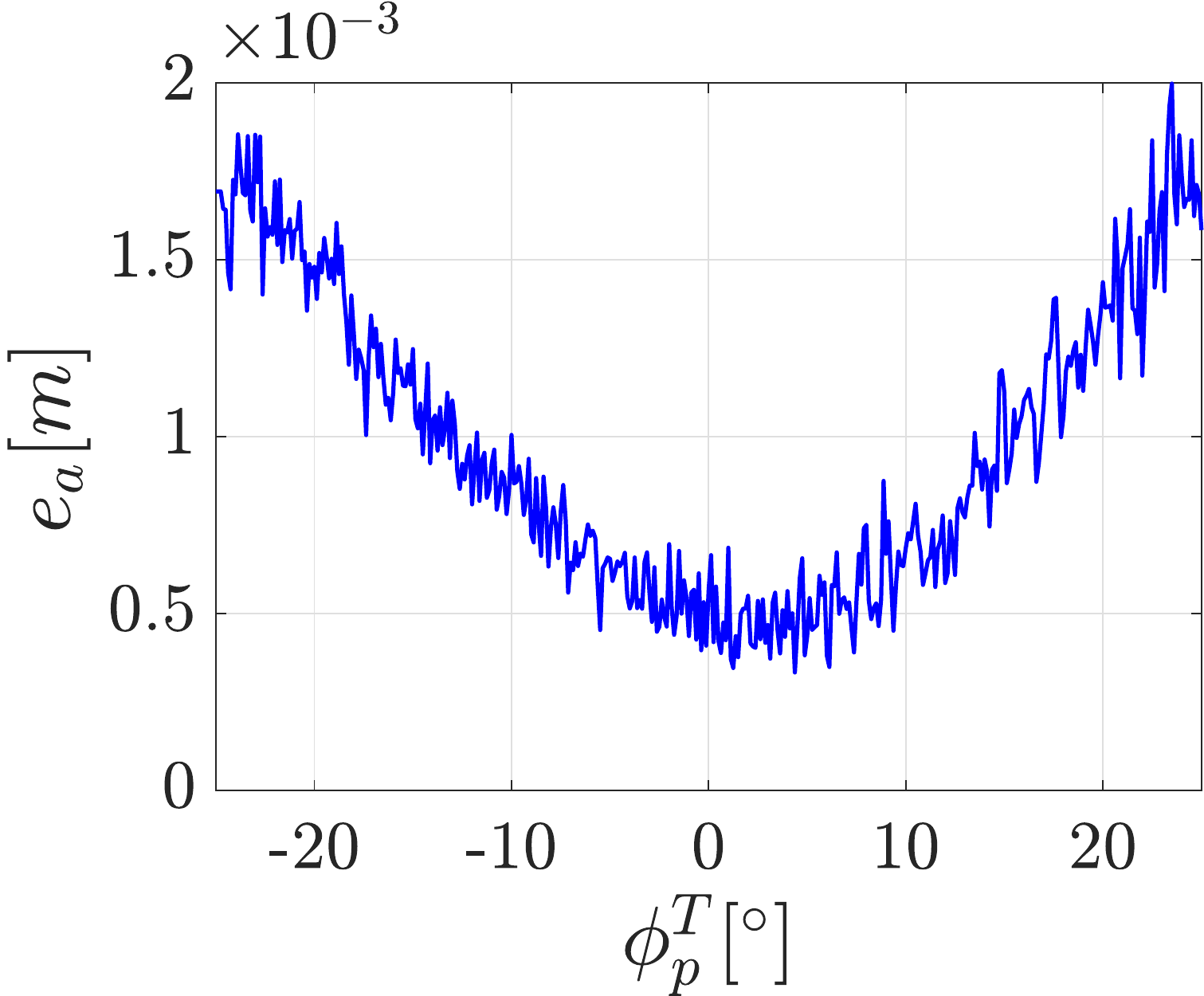}
\includegraphics[width=0.49\linewidth]{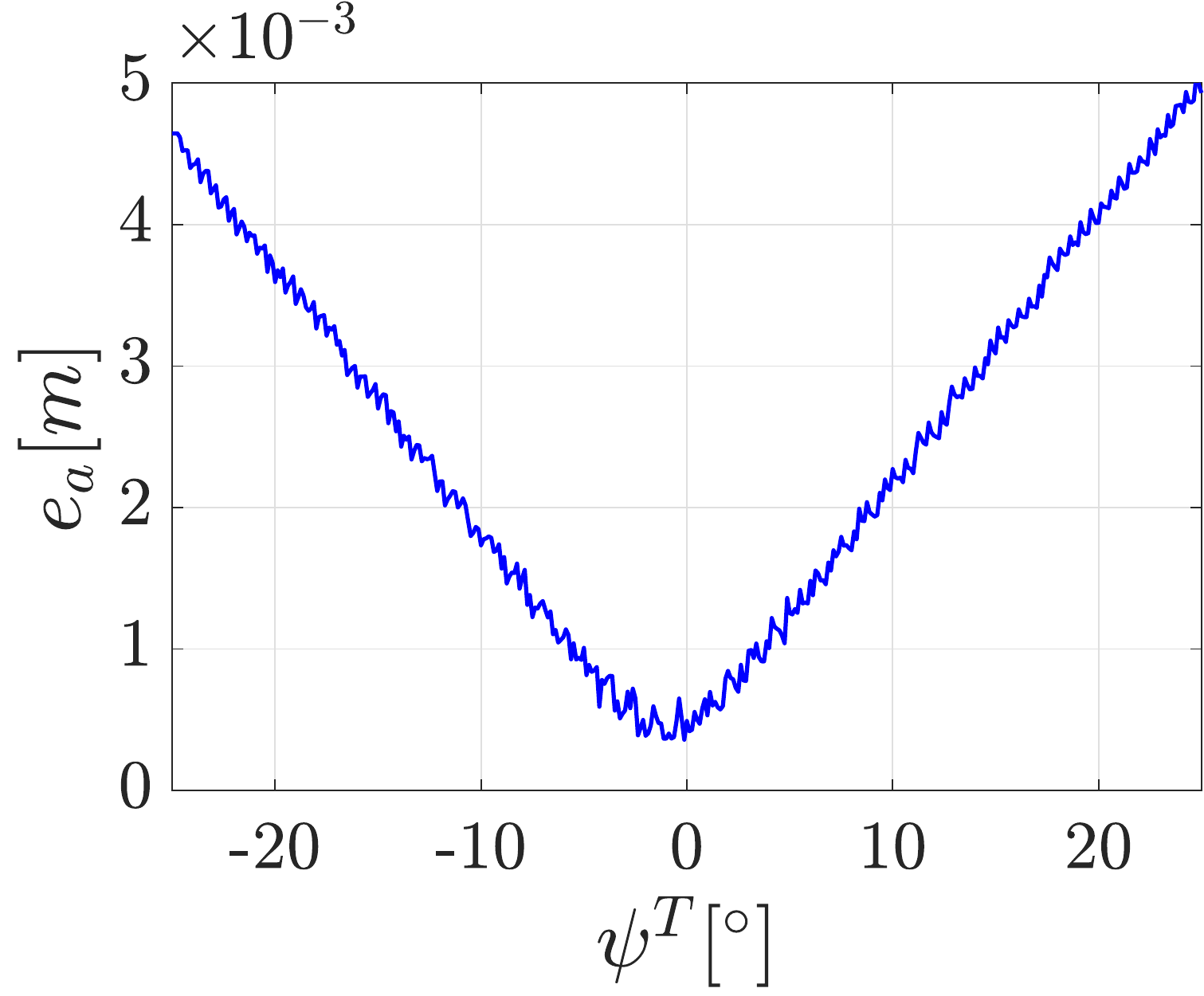}}
\caption{Accuracy of the presented triangle approximation, with varying pendulum extension and trunk orientation parameters.}
\figlabel{com_accuracy}
\vspace{-2.5ex}
\end{figure}

To confirm that the assumptions made are correct, we measure the error $e_a$ between
the CoM calculated with the URDF model (treated as ground truth) and the CoM 
calculated using our triangle approximation. A set of motions has been generated using the proposed
method to assess the influence of the most critical parameters on the accuracy---the pendulum extensions
and trunk orientation. The first motion alters the body pendulum length $l^B$ across the whole possible range---
from \SI{0.25}{m} to \SI{0.41}{m} with the trunk kept upright. The next motions
keep $l^B$ constant at \SI{0.30}{m}, and vary the trunk orientation about each axis separately. 
The result of the approximation can be seen in \figref{com_accuracy}.

In a typical scenario where the trunk is upright, the error is generally below \SI{1}{mm} and
can reach as low as \SI{0.2}{mm}. The biggest contribution to the decrease in accuracy has the trunk orientation. 
As the trunk is the root part that connects all four of the limbs, their movement is paired 
with any change in the trunks orientation. As that movement breaks the symmetry, any error becomes quickly visible.
The mass distribution parameters then become essential to proper CoM calculation and limb placement.
This can be observed best with a change in the trunk projected pitch $\theta^T_p$ and fused yaw $\psi^T$ angles.
The trunk projected roll $\phi^T_p$, also shows this property, but to a lesser extent.
When the trunk is upright, all of the masses generally lie on the frontal plane and the error is mostly two-dimensional.
Pitching or yawing the trunk moves the masses further away from the plane, increasing the errors.
The error is symmetrical and grows linearly, which leads to suggest that it is an effect
of a slight error in the calibration of the mass distribution parameters. 

As the calibration was done with a single data point, the results can still be improved. 
One possibility of addressing this would involve averaging several data points at different 
joint configurations. Searching for an optimal set of parameters with an error minimising function is also feasible. 
In spite of the minor miscalibration, the errors in the possible range of movement do not exceed \SI{5}{mm},
which translates to a maximum error of \SI{1.6}{\%}. The achieved results are considered to be
sufficiently accurate in order to apply the method on a real robot. 

\subsection{Balanced whole body posing}

\begin{figure}[!t]
\parbox{\linewidth}{\centering
\includegraphics[height=0.30\linewidth]{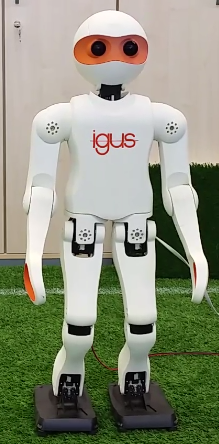}\hspace{0.05px}
\includegraphics[height=0.30\linewidth]{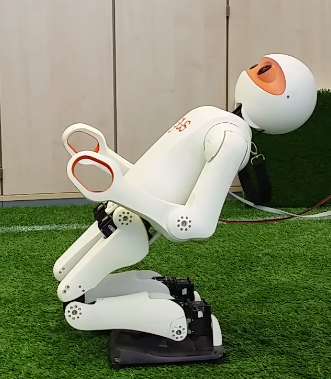}\hspace{0.05px}
\includegraphics[height=0.30\linewidth]{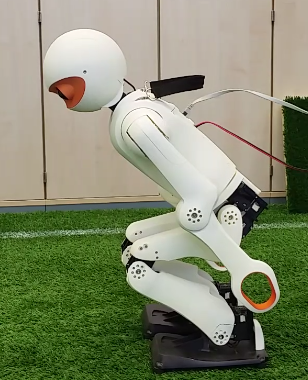}\hspace{0.05px}
\includegraphics[height=0.30\linewidth]{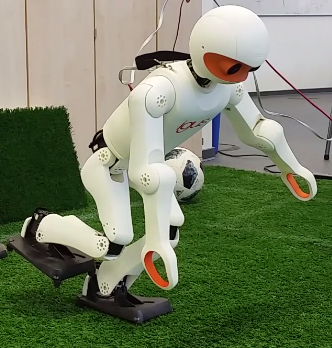}}
\caption{The \iguhop in several balanced motion keyframes, with varying pendulum extension, trunk orientation, and support coefficient values.}
\figlabel{robot_poses}
\vspace{-1ex}
\end{figure}

We verify our motion generator, with multiple test poses in 
single and double leg support with various trunk orientations; a sample of 
these can be observed in \figref{robot_poses}. The motions to reach these poses
were generated at runtime by modifying the parameters introduced at the beginning of \secref{pose_generation}.

One noteworthy property to mention is that with the trunk and pendulum fully 
upright and a nominal pendulum extension, our method generates a natural looking
standby pose, as seen on the left of \figref{robot_poses}. On the same figure it can be 
observed that even quite complex motions in terms of balance can be performed. This is 
a result of the inherent vertical orientation of the body pendulum when generating the pose.

\subsection{Kicking motion}

\begin{figure}[!t]
\parbox{\linewidth}{\centering
\includegraphics[height=0.25\linewidth]{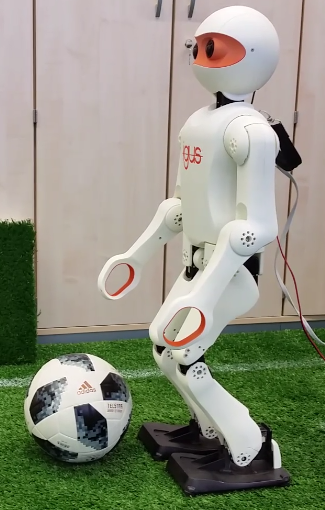}\hspace{0.05px}
\includegraphics[height=0.25\linewidth]{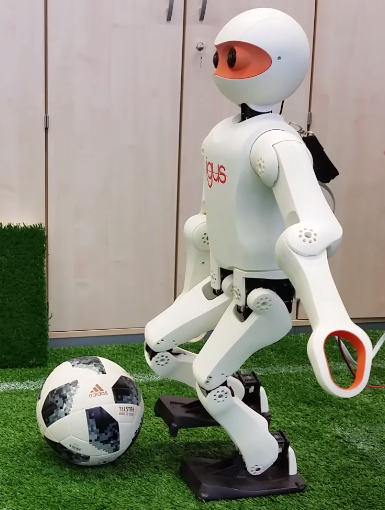}\hspace{0.05px}
\includegraphics[height=0.25\linewidth]{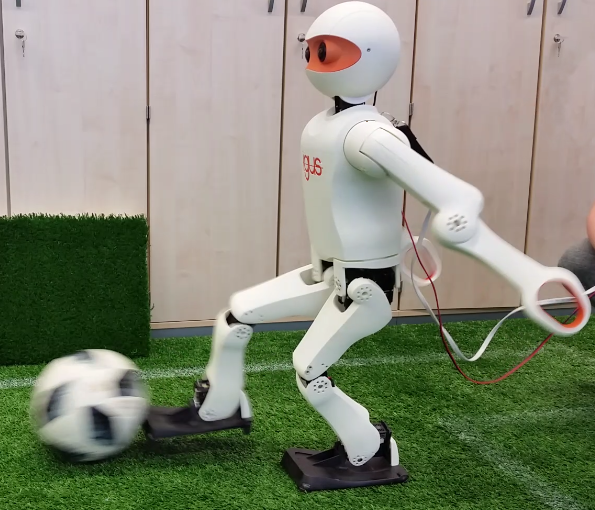}\hspace{0.05px}
\includegraphics[height=0.25\linewidth]{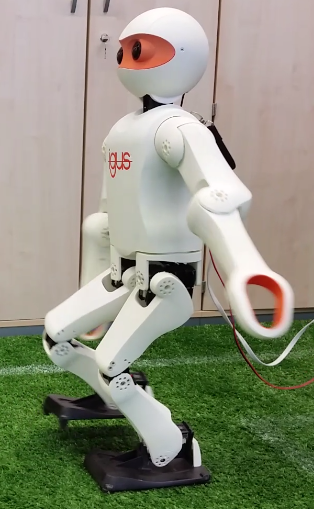}\hspace{0.05px}
\includegraphics[height=0.25\linewidth]{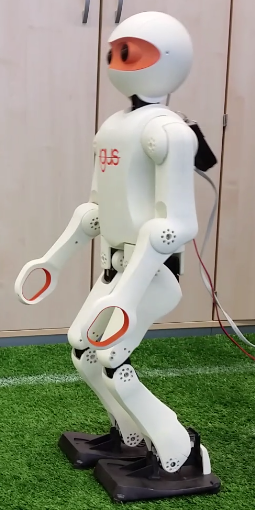}}
\caption{Still frames from the generated kicking motions.}
\figlabel{wbkick}
\vspace{-1ex}
\end{figure}

Generating a motion to achieve a balanced pose which is parametric is a very useful 
feature in itself and leads to performing balanced task-oriented motions.
An example of this is a kicking motion, where shifting the whole weight of the body
onto a single leg, relieves the other one to perform the kick. The trajectory of 
the foot can then be computed as to hit the ball for a specific result~(e.g. side kicks,
high kicks, pass kicks and so on). For this motion, only the support 
coefficient $c_s$ value, and desired foot position $F^R$ were changed through time. 
The achieved kick~(seen~on~\figref{wbkick}) is relatively dynamic. The fastest transitions 
between frames are done in \SI{0.15}{s}.

\begin{figure}[!t]
\parbox{\linewidth}{\centering\includegraphics[width=0.99\linewidth]{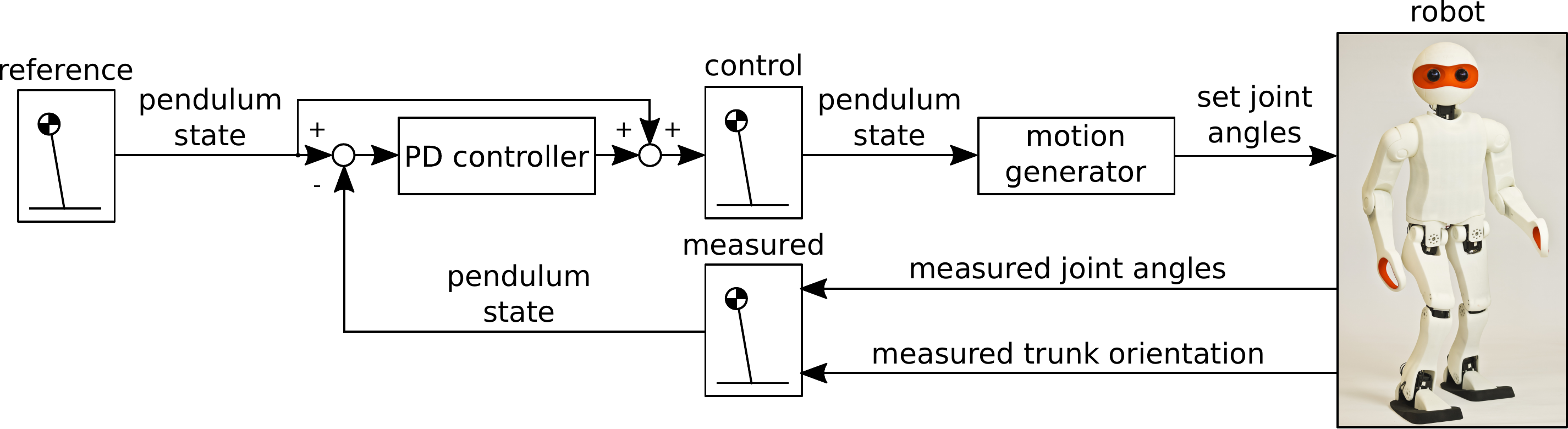}}
\caption{Control schematic depicting the PD stabilisation. Measurements from the robot are used to 
calculate an offset pendulum state through PD control on three variables: $\phi_{p}^B$, $\theta_{p}^B$ and $l^B$. 
The resulting pendulum state is used as the input for the motion generator.}
\figlabel{schematic}
\vspace{-3ex}
\end{figure}

Due to the low torque-to-weight ratio in the \iguhop, the actuators on the robot generally experience difficulties 
following their reference position, which leads to large CoM tracking errors. This is particularly
visible in the ankle joint of the supporting foot. Although the robot can tolerate this to some extent, 
the precision is decreased. As the motions are generated on-the-fly, we have the possibility 
to modify them with several corrective actions in response to the CoM tracking error. One of the 
possibilities is to generate motions with an offset in the desired pendulum projected roll $\phi_{p}^B$ 
and pitch $\theta_{p}^B$, as well as compensating for the pendulum extension $l^B$. For 
simplicity, we generate the corrective poses through separate PD regulators on the mentioned
variables. A schematic showing the utilised approach can be seen on \figref{schematic}.

\begin{figure*}[!t]
\centering\parbox{0.9\linewidth}{\centering
\includegraphics[width=0.49\linewidth]{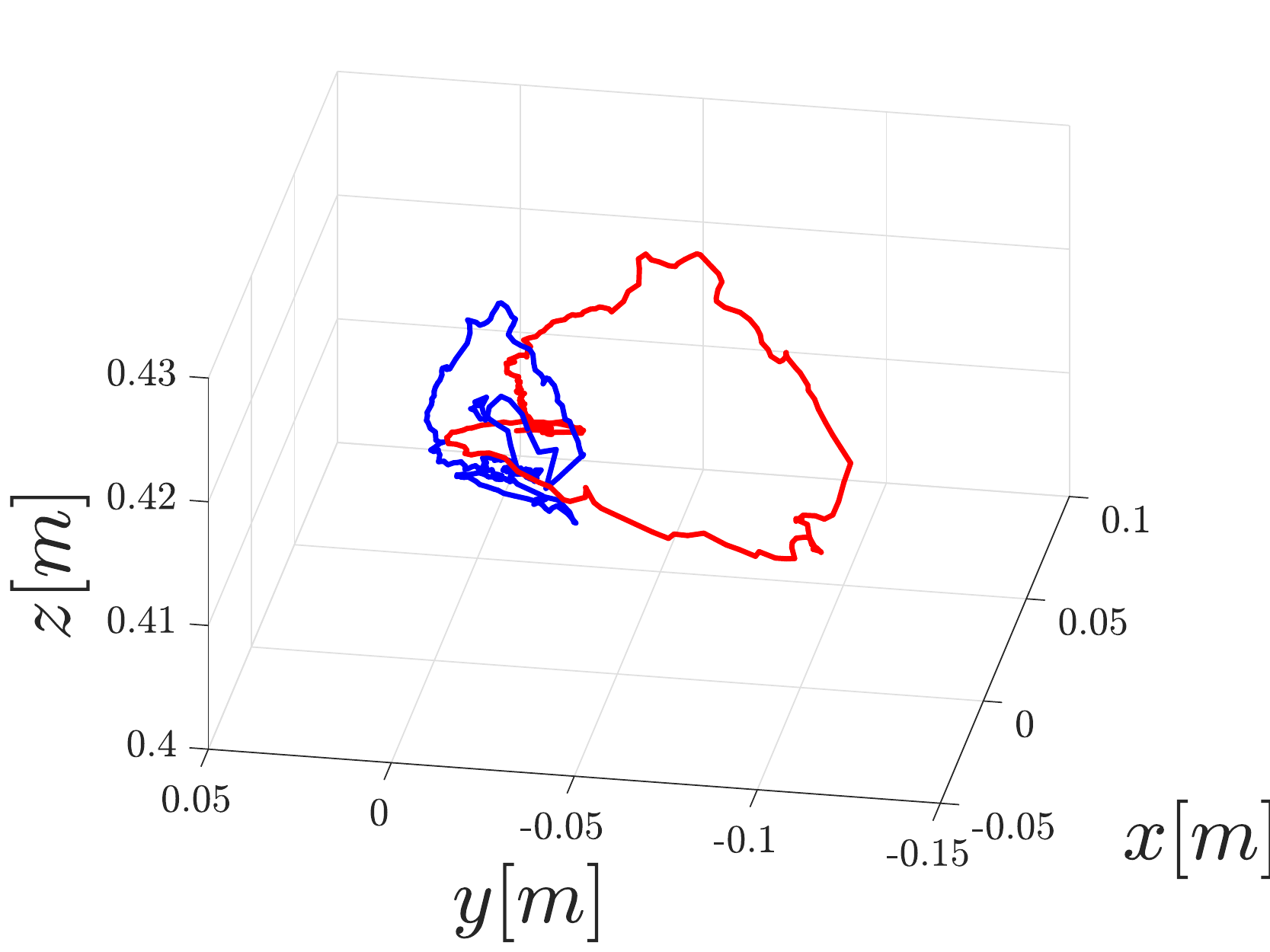}
\includegraphics[width=0.49\linewidth]{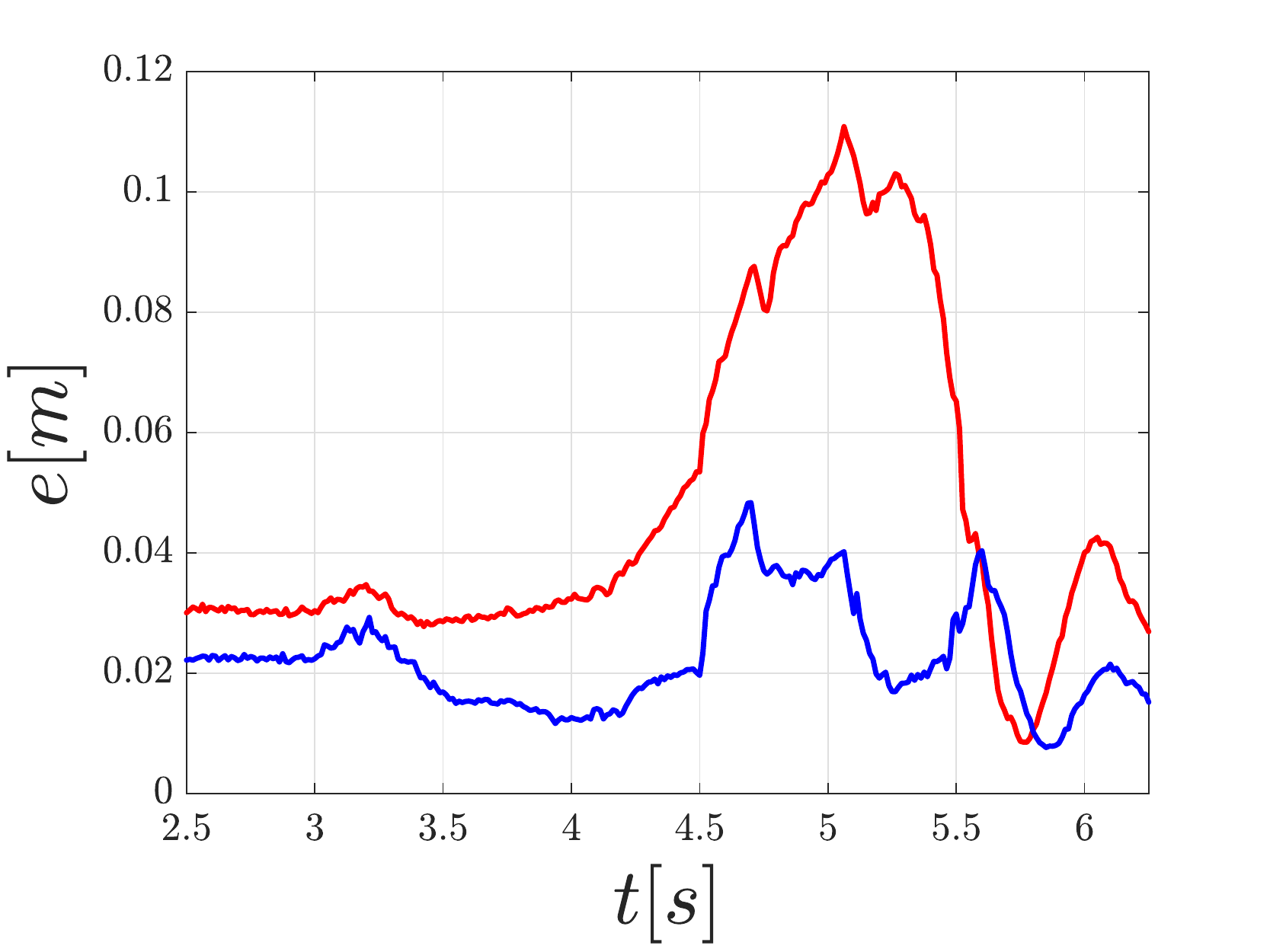}}
\caption{measured CoM movement~(left) and tracking error~(right) while kicking. Red represents the open-loop motion, while blue shows the same motion with PD stabilisation enabled. }
\figlabel{com_kick}
\vspace{-3ex}
\end{figure*}

The achieved result in the form of measured CoM with and without PD stabilisation during the kicking motion is shown
in \figref{com_kick}. When performing the motion open-loop, the steady-state error was in the range of \SI{3}{cm}.
After lifting the foot, the robot had a tendency to lean towards the kicking foot, and the disturbance of 
making contact between the foot and the ball led to CoM tracking errors of almost \SI{12}{cm}. 
These errors can be attributed to the actuation limitations of the \iguhop, where the ankle actuators are underpowered. After applying PD stabilisation on the reference
upright-oriented pendulum, the movement of the CoM was much more contained and symmetric. On average, the error was in the range of \SI{2}{cm}
and did not exceed \SI{5}{cm}. The steady-state error also decreased. Applying the feedback mechanism also allowed the robot to
reject a moderate level of disturbances, which was not possible otherwise.

\section{Conclusions}

In this work, we presented an analytic, geometric motion generation method for humanoid robots, 
based on body mass distribution. The statically balanced motions are produced from a small set of directly 
comprehensible parameters in a frame-by-frame manner at interactive rates. As the approach does not rely on 
any force or torque information, it is applicable to platforms which lack the necessary sensing 
and actuation capabilities for state-of-the-art Whole-Body Control algorithms. 
We have experimentally verified the method on the \iguhop and were able to produce stable kicking motions
which was previously not possible without meticulous design and extensive tuning. 
One of the most notable merits of the approach is its extendability, which we demonstrated by adding simple
PD stabilisation mechanisms on the body pendulum. In future work, feedback mechanisms can be further extended and involve 
for example, changing the trunk and limb placement strategies. As a pendulum is an integral part of the approach,
a whole class of model-based control techniques can be applied to control the robot. The motion generator then is used to tie
the controlled pendulum state with a whole-body pose, ensuring that the CoM is accurately tracked.
This makes the proposed motion generator a powerful building block for more sophisticated control schemes.

%

%
%

%
%

%

%
%
%

%
%
%
%
%
%
%

%
%
%
%
%

%
%
%
%
%
%
%
%

%
%
%

%

%
%
%
%

%
%
\bibliographystyle{IEEEtran}
\bibliography{IEEEabrv,ms}
\balance

\end{document}

%% file: mass_distribution_triangle.pdf_tex
\begingroup%
  \makeatletter%
  \providecommand\color[2][]{%
    \errmessage{(Inkscape) Color is used for the text in Inkscape, but the package 'color.sty' is not loaded}%
    \renewcommand\color[2][]{}%
  }%
  \providecommand\transparent[1]{%
    \errmessage{(Inkscape) Transparency is used (non-zero) for the text in Inkscape, but the package 'transparent.sty' is not loaded}%
    \renewcommand\transparent[1]{}%
  }%
  \providecommand\rotatebox[2]{#2}%
  \newcommand*\fsize{\dimexpr\f@size pt\relax}%
  \newcommand*\lineheight[1]{\fontsize{\fsize}{#1\fsize}\selectfont}%
  \ifx\svgwidth\undefined%
    \setlength{\unitlength}{299.56617055bp}%
    \ifx\svgscale\undefined%
      \relax%
    \else%
      \setlength{\unitlength}{\unitlength * \real{\svgscale}}%
    \fi%
  \else%
    \setlength{\unitlength}{\svgwidth}%
  \fi%
  \global\let\svgwidth\undefined%
  \global\let\svgscale\undefined%
  \makeatother%
  \begin{picture}(1,0.78847272)%
    \lineheight{1}%
    \setlength\tabcolsep{0pt}%
    \put(0,0){\includegraphics[width=\unitlength,page=1]{mass_distribution_triangle.pdf}}%
    \put(-0.00172124,0.00625905){\color[rgb]{0,0,0}\makebox(0,0)[lt]{\lineheight{1.25}\smash{\begin{tabular}[t]{l}$A$\end{tabular}}}}%
    \put(0.17278249,0.7472673){\color[rgb]{0,0,0}\makebox(0,0)[lt]{\lineheight{1.25}\smash{\begin{tabular}[t]{l}$B$\end{tabular}}}}%
    \put(0.85562278,0.33376941){\color[rgb]{0,0,0}\makebox(0,0)[lt]{\lineheight{1.25}\smash{\begin{tabular}[t]{l}$C$\end{tabular}}}}%
    \put(0.17731695,0.33956833){\color[rgb]{0,0,0}\makebox(0,0)[lt]{\lineheight{1.25}\smash{\begin{tabular}[t]{l}$M$\end{tabular}}}}%
    \put(0.41609404,0.62020172){\color[rgb]{0,0,0}\makebox(0,0)[lt]{\lineheight{1.25}\smash{\begin{tabular}[t]{l}$P$\end{tabular}}}}%
    \put(0.08052511,0.40946539){\color[rgb]{0,0,0}\makebox(0,0)[lt]{\lineheight{1.25}\smash{\begin{tabular}[t]{l}$c$\end{tabular}}}}%
    \put(0.46358479,0.15244574){\color[rgb]{0,0,0}\makebox(0,0)[lt]{\lineheight{1.25}\smash{\begin{tabular}[t]{l}$b$\end{tabular}}}}%
    \put(0,0){\includegraphics[width=\unitlength,page=2]{mass_distribution_triangle.pdf}}%
    \put(0.20945352,0.66254452){\color[rgb]{0,0,0}\makebox(0,0)[lt]{\lineheight{1.25}\smash{\begin{tabular}[t]{l}$\beta$\end{tabular}}}}%
    \put(0.74940324,0.34767923){\color[rgb]{0,0,0}\makebox(0,0)[lt]{\lineheight{1.25}\smash{\begin{tabular}[t]{l}$\gamma$\end{tabular}}}}%
    \put(0.33818382,0.58640907){\color[rgb]{0,0,0}\makebox(0,0)[lt]{\lineheight{1.25}\smash{\begin{tabular}[t]{l}$\gamma_1$\end{tabular}}}}%
    \put(0.4103746,0.53541185){\color[rgb]{0,0,0}\makebox(0,0)[lt]{\lineheight{1.25}\smash{\begin{tabular}[t]{l}$\beta_2$\end{tabular}}}}%
    \put(0.30087345,0.39438445){\color[rgb]{0,0,1}\makebox(0,0)[lt]{\lineheight{1.25}\smash{\begin{tabular}[t]{l}$(1-p_l)l$\end{tabular}}}}%
    \put(0.16004119,0.17175091){\color[rgb]{1,0.63137255,0}\makebox(0,0)[lt]{\lineheight{1.25}\smash{\begin{tabular}[t]{l}$p_ll$\end{tabular}}}}%
    \put(0.58491963,0.52075823){\color[rgb]{0.16470588,0.67843137,0.16470588}\makebox(0,0)[lt]{\lineheight{1.25}\smash{\begin{tabular}[t]{l}$(1-p_s)a$\end{tabular}}}}%
    \put(0.29408022,0.70790706){\color[rgb]{1,0,0}\makebox(0,0)[lt]{\lineheight{1.25}\smash{\begin{tabular}[t]{l}$p_sa$\end{tabular}}}}%
    \put(0,0){\includegraphics[width=\unitlength,page=3]{mass_distribution_triangle.pdf}}%
    \put(0.07036455,0.08779657){\color[rgb]{0,0,0}\makebox(0,0)[lt]{\lineheight{1.25}\smash{\begin{tabular}[t]{l}$\alpha$\end{tabular}}}}%
    \put(0,0){\includegraphics[width=\unitlength,page=4]{mass_distribution_triangle.pdf}}%
  \end{picture}%
\endgroup%

%% file: projected_angles.pdf_tex
\begingroup%
  \makeatletter%
  \providecommand\color[2][]{%
    \errmessage{(Inkscape) Color is used for the text in Inkscape, but the package 'color.sty' is not loaded}%
    \renewcommand\color[2][]{}%
  }%
  \providecommand\transparent[1]{%
    \errmessage{(Inkscape) Transparency is used (non-zero) for the text in Inkscape, but the package 'transparent.sty' is not loaded}%
    \renewcommand\transparent[1]{}%
  }%
  \providecommand\rotatebox[2]{#2}%
  \newcommand*\fsize{\dimexpr\f@size pt\relax}%
  \newcommand*\lineheight[1]{\fontsize{\fsize}{#1\fsize}\selectfont}%
  \ifx\svgwidth\undefined%
    \setlength{\unitlength}{263.12375717bp}%
    \ifx\svgscale\undefined%
      \relax%
    \else%
      \setlength{\unitlength}{\unitlength * \real{\svgscale}}%
    \fi%
  \else%
    \setlength{\unitlength}{\svgwidth}%
  \fi%
  \global\let\svgwidth\undefined%
  \global\let\svgscale\undefined%
  \makeatother%
  \begin{picture}(1,0.7330749)%
    \lineheight{1}%
    \setlength\tabcolsep{0pt}%
    \put(0,0){\includegraphics[width=\unitlength,page=1]{projected_angles.pdf}}%
    \put(0.36171317,0.38951862){\color[rgb]{0.09019608,0.96470588,0.94901961}\makebox(0,0)[lt]{\lineheight{1.25}\smash{\begin{tabular}[t]{l}$\theta_p$\end{tabular}}}}%
    \put(0.51475535,0.7002784){\color[rgb]{0,0,0.96470588}\makebox(0,0)[lt]{\lineheight{1.25}\smash{\begin{tabular}[t]{l}$z_G$\end{tabular}}}}%
    \put(0.04773738,0.12746996){\color[rgb]{1,0,0}\makebox(0,0)[lt]{\lineheight{1.25}\smash{\begin{tabular}[t]{l}$x_G$\end{tabular}}}}%
    \put(0.87611409,0.11835509){\color[rgb]{0.16470588,0.67843137,0.16470588}\makebox(0,0)[lt]{\lineheight{1.25}\smash{\begin{tabular}[t]{l}$y_G$\end{tabular}}}}%
    \put(0.55821314,0.33999145){\color[rgb]{0,0,0}\makebox(0,0)[lt]{\lineheight{1.25}\smash{\begin{tabular}[t]{l}$z_B$\end{tabular}}}}%
    \put(0.02456489,0.44649498){\color[rgb]{0.16470588,0.67843137,0.16470588}\rotatebox{14.248494}{\makebox(0,0)[lt]{\lineheight{1.25}\smash{\begin{tabular}[t]{l}$x_Gz_G$\end{tabular}}}}}%
    \put(0.83104215,0.47393476){\color[rgb]{1,0,0}\rotatebox{-14.433048}{\makebox(0,0)[lt]{\lineheight{1.25}\smash{\begin{tabular}[t]{l}$y_Gz_G$\end{tabular}}}}}%
    \put(0.55948416,0.4488911){\color[rgb]{0.85098039,0.10980392,0.71764706}\makebox(0,0)[lt]{\lineheight{1.25}\smash{\begin{tabular}[t]{l}$\phi_p$\end{tabular}}}}%
    \put(0.80760074,0.27855244){\color[rgb]{0,0,0}\makebox(0,0)[lt]{\lineheight{1.25}\smash{\begin{tabular}[t]{l}$z_{B_z}$\end{tabular}}}}%
    \put(0.60391755,0.54290706){\color[rgb]{0,0,0}\makebox(0,0)[lt]{\lineheight{1.25}\smash{\begin{tabular}[t]{l}$z_{B_y}$\end{tabular}}}}%
    \put(0,0){\includegraphics[width=\unitlength,page=2]{projected_angles.pdf}}%
    \put(0.31661252,0.54711186){\color[rgb]{0,0,0}\makebox(0,0)[lt]{\lineheight{1.25}\smash{\begin{tabular}[t]{l}$z_{B_x}$\end{tabular}}}}%
    \put(0.16750234,0.2975688){\color[rgb]{0,0,0}\makebox(0,0)[lt]{\lineheight{1.25}\smash{\begin{tabular}[t]{l}$z_{B_z}$\end{tabular}}}}%
    \put(0,0){\includegraphics[width=\unitlength,page=3]{projected_angles.pdf}}%
  \end{picture}%
\endgroup%